\documentclass[a4paper,12pt,reqno]{amsart}
\usepackage{amssymb}
\usepackage{amsmath}
\usepackage{mathtools}
\usepackage{enumitem}
\usepackage{mathrsfs}
\usepackage[all]{xy}
\usepackage{mathtools}
\usepackage{hyperref}
\usepackage{url}
\usepackage{bbm}
\usepackage{longtable}
\usepackage{booktabs}

\usepackage[OT2,T1]{fontenc}
\DeclareSymbolFont{cyrletters}{OT2}{wncyr}{m}{n}
\usepackage[british]{babel}
\numberwithin{equation}{section} \numberwithin{figure}{section}

\usepackage[margin=1in]{geometry}
\usepackage[dvipsnames]{xcolor}

\renewcommand{\paragraph}[1]{\vspace*{1mm} \noindent \textbf{#1} \hspace*{1mm}}

\newcommand\ZZ{\mathbb{Z}}

\newcommand\Z{\mathbb{Z}}

\newtheorem{lemma}{Lemma}

\theoremstyle{definition}
\newtheorem{example}[lemma]{Example}
\newtheorem{definition}[lemma]{Definition}

\newtheorem*{ack}{Acknowledgments}
\makeatletter
\let\@mkboth\@gobbletwo
\let\@evenhead\@oddhead
\makeatother

\title{Machine Learning for Modular Multiplication}
\address{$^{\ast}$Authors in alphabetical order}

\author{Kristin Lauter$^{\ast, 1} $}
\address{1.\ Meta AI \\ \href{mailto:klauter@meta.com}{klauter@meta.com}} 
\author{Cathy Yuanchen Li$^{\ast, 2}$}
\address{2.\ University of Chicago  \\ \href{mailto:yuanchen@cs.uchicago.edu}{yuanchen@cs.uchicago.edu}}
\author{Krystal Maughan$^{\ast, 3}$ }
\address{3.\ University of Vermont \\ \href{mailto:Krystal.Maughan@uvm.edu}{Krystal.Maughan@uvm.edu}}
\author{Rachel Newton$^{\ast, 4}$ } \address{4.\ King's College London \\ \href{mailto:rachel.newton@kcl.ac.uk}{rachel.newton@kcl.ac.uk}}
\author{Megha Srivastava$^{\ast, 5}$ } \address{5.\ Stanford University  \\ \href{mailto:megha@cs.stanford.edu}{megha@cs.stanford.edu}}

\usepackage[foot]{amsaddr}

\date{}

\let\origmaketitle\maketitle
\def\maketitle{
  \begingroup
  \let\MakeUppercase\relax 
  \origmaketitle
  \endgroup
}

\begin{document}

\maketitle


\begin{abstract}
Motivated by cryptographic applications, we investigate two machine learning approaches to modular multiplication: namely circular regression and a sequence-to-sequence transformer model. The limited success of both methods demonstrated in our results gives evidence for the hardness of tasks involving modular multiplication upon which cryptosystems are based.
\end{abstract}
\section{Introduction}

Machine learning approaches to modular arithmetic have recently entered the limelight, motivated at least in part by the reliance on modular addition and multiplication of several cryptographic schemes, including RSA, Learning With Errors (LWE) and Diffie--Hellman, to name but a few. 

In the SALSA series of papers \cite{wenger2022salsa, li2023salsapicante, verde}, transformers are used to develop novel machine learning attacks on LWE-based cryptographic schemes. Such schemes are based on the following problem, which is assumed to be hard. Given a dimension $n$, an integer modulus $q$, and a secret vector $s\in\ZZ_q^n$, solving the Learning With Errors problem~\cite{Regev} entails finding the secret $s$ given a data set consisting of pairs of vectors $(a_i,b_i)$ where $a_i\in\ZZ_q^n$ and $b_i\equiv a_i\cdot s+e_i\pmod{q}$ with $e_i$ a small `error' or `noise' sampled from a narrow centered Gaussian distribution. We call $b_i$ a `noisy inner product'. In~\cite{wenger2022salsa, li2023salsapicante, verde}, transformer models are (partially) trained on noisy inner product data and used along with statistical cryptanalysis techniques to recover the secret $s$, under certain conditions on the dimension, modulus and secret size/sparseness. 

In dimension $n=1$, recovering the secret $s$ is akin to finding a modular inverse for $a_i$. Therefore, it is natural to ask whether the SALSA architecture has in fact learnt modular arithmetic operations such as multiplication and division. However, the latest paper in the series, SALSA VERDE~\cite{verde}, calls this into question. The authors of that paper observe that secret recovery is harder when the modulus $q$ is smaller and suggest that this may be a modular arithmetic issue. When $q$ is large compared to $a_i$, $s$ and $e_i$, computing $b$ does not involve any modular arithmetic (modular multiplication is replaced by ordinary multiplication). This reopens the question of whether transformers or other machine learning techniques can tackle modular multiplication and provides motivation for our investigations in the present paper. 

We experiment with two different machine learning approaches to modular multiplication. In Section~\ref{sec:circreg} we describe an approach to solving the $1$-dimensional Learning With Errors problem using circular regression. We show that the method, as implemented, does not consistently give a significant improvement on exhaustive search. Indeed, some recent theoretical results suggest that in general a large number of iterations are needed in order for methods based on gradient descent to successfully tackle modular multiplication~\cite{takhanov23}.
In Section~\ref{sec:transformer} we explore an alternative approach using a sequence-to-sequence transformer, and show poor generalization performance of our method for several different values of a secret $s$ and base representations of numbers, likewise demonstrating the difficult nature of this problem, in line with evidence presented in~\cite{Palamas, Jelassi23}. We end with a discussion of the relevance of modular arithmetic to cryptographic schemes such as Diffie--Hellman in Section~\ref{sec:discuss}.

\begin{ack}
 We are grateful to Julio Brau and Kolyan Ray for useful discussions.
  Rachel Newton was supported by UKRI Future Leaders Fellowship MR/T041609/1 and MR/T041609/2. Megha Srivastava was supported by an IBM PhD Fellowship and the NSF Graduate Research Fellowship Program under Grant No.~DGE-1656518. This project began at the WIN6: Women in Numbers 6 workshop in 2023.
  We are grateful for the hospitality and support of the Banff International Research Station during the workshop and we would especially like to thank
  Shabnam Akhtari, Alina Bucur, Jennifer Park and
Renate Scheidler for organizing WIN6 and making this collaboration possible.
\end{ack}

\section{Circular regression for modular multiplication}\label{sec:circreg}
\subsection{The task}\label{circular-task} We focus on the 1-dimensional version of LWE.
Let $p\in\mathbb{Z}_{>0}$ be a prime number and let $s\in\Z/p\Z$. We will refer to $s$ as the \emph{secret}. Given a data set consisting of pairs of integers $\{(a_i,b_i)\}_{1\leq i\leq m}$, where $b_i\equiv a_is+e_i\pmod{p}$ and the `noise' or `error' values $e_i$ are sampled from a centered discrete Gaussian distribution with standard deviation $\sigma$, the task is to find the unknown secret $s$. This problem was studied in the case of binary secrets in~\cite{circular}, where the authors explored using circular regression to solve Learning With Errors (LWE) in small dimension up to $28$.

\subsection{Transforming to a circular regression problem} 
Following~\cite{circular}, we rescale to view our integers modulo $p$ as points on the unit circle: define $y_i=\frac{2\pi}{p}b_i$ so that the congruence $b_i\equiv a_is+e_i\pmod{p}$ becomes 
\[y_i\equiv \frac{2\pi}{p}a_is+\frac{2\pi}{p}e_i\pmod{2\pi}.\]
As in~\cite{circular}, we assume the target variables $y_i$ follow a discrete von Mises distribution. The von Mises distribution is also known as circular normal distribution or Tikhonov distribution and it closely approximates a wrapped normal distribution. 

\begin{definition}[von Mises distribution]
The von Mises distribution is a continuous probability distribution on the circle. The von Mises probability density function for the angle $\theta$ is
\[ f(\theta \mid \mu ,\kappa )={\frac {\exp(\kappa \cos(\theta-\mu ))}{2\pi I_{0}(\kappa )}}\]
where the parameters $\mu$ and $1/\kappa$ are analogous to $\mu$ and $\sigma^2$ (the mean and variance) in the normal distribution:
\begin{itemize}
    \item $\mu$ is a measure of location (the distribution is clustered around $\mu$), and
    \item $\kappa$  is a measure of concentration (a reciprocal measure of dispersion, so $1/\kappa$ is analogous to $\sigma^2$).
\end{itemize}
$I_{0}(\kappa )$ is a scaling constant chosen so that the distribution is a probability distribution, i.e.\ it sums to $1$. $I_{0}$ denotes the modified Bessel function of the first kind of order 0, which satisfies
\[ \int _{-\pi }^{\pi }\exp(\kappa \cos x)dx={2\pi I_{0}(\kappa )}.\]
\end{definition}

Since our samples $y_i$ correspond to the integers $b_i$, they can only take certain discrete values, namely those angles of the form $2\pi n/p$ for integers $n\in [\frac{-p+1}{2},\frac{p-1}{2}]$. To build a discrete version of the von Mises distribution, take a parameter $c$ and define a modified probability distribution function $\tilde{f}_c(\theta \mid \mu ,\kappa )$ as follows:
\[ \tilde{f}_c(\theta \mid \mu ,\kappa )=\begin{cases}
   c \cdot f(\theta \mid \mu ,\kappa )& \textrm{ if } \theta=2\pi n/p \textrm{ for some } n\in\mathbb{Z} \cap [\frac{-p+1}{2},\frac{p-1}{2}]; \\
    0 & \textrm{ otherwise}.
\end{cases}\]
To make this a probability distribution, we need the distribution to sum to $1$; this is achieved by choosing an appropriate value for $c$. Assume from now on that such a value $c$ has been fixed. 

We consider each $y_i$ as being sampled from a discrete von Mises distribution $\tilde{f}_c(\theta \mid \mu_i ,\kappa)$ where $\mu_i=\frac{2\pi}{p}a_is$ and $\kappa$ is the same for each $b_i$ since it corresponds to the variance $\sigma^2$ of the distribution of the errors $e_i$.

The events $y_i$ are independent so the probability density function for a collection of samples $y_1, \dots , y_m$ is
\begin{align*}
    \prod_{i=1}^m \tilde{f}_c(y_i \mid \mu_i ,\kappa)=c^m \prod_{i=1}^m f(y_i \mid \mu_i ,\kappa).
\end{align*}

Therefore, the likelihood function for the collection of samples, given parameters \\
$(\mu_1,\dots , \mu_m; \kappa)=(\frac{2\pi}{p}a_1s,\dots , \frac{2\pi}{p}a_ms; \kappa)$ is 
\begin{align*}
  \mathcal{L}((\mu_1,\dots , \mu_m; \kappa)\mid (y_1,\dots, y_m))  =c^m \prod_{i=1}^m f(y_i \mid \mu_i ,\kappa).
\end{align*}

Since $c$ is fixed, maximizing the (log-)likelihood function is equivalent to maximising 
\begin{align*}
    \sum_{i=1}^m \log f(y_i \mid \mu_i ,\kappa)=\kappa \sum_{i=1}^m\cos\left(y_i-\frac{2\pi}{p}a_is \right) -m\log (2\pi I_0(\kappa)).
\end{align*}

Thus, we seek to maximize $\sum_{i=1}^m\cos\left(y_i-\frac{2\pi}{p}a_is \right)$ over all choices of $s$. Equivalently, we minimise the \emph{circular regression loss} $-\sum_{i=1}^m\cos\left(y_i-\frac{2\pi}{p}a_is \right)$. In pursuit of this goal, we will treat $s$ as if it were a continuous real variable, use gradient descent to find a real value of $s$ that minimizes $-\sum_{i=1}^m\cos\left(y_i-\frac{2\pi}{p}a_is \right)$, and our final output will be the integer closest to this real value of $s$.

Differentiating $-\sum_{i=1}^m\cos\left(y_i-\frac{2\pi}{p}a_is \right)$ with respect to $s$ gives the \emph{gradient}
\begin{align*}
    -\frac{2\pi}{p}\sum_{i=1}^m a_i\sin\left(y_i-\frac{2\pi}{p}a_is \right).
\end{align*}
So we start with an initial guess for $s$, call it $s_0$, and at each time step $t$ we define
\begin{align*}
    s_{t+1}=s_t+\eta\frac{2\pi}{p}\sum_{i=1}^m a_i\sin\left(y_i-\frac{2\pi}{p}a_is_t \right),
\end{align*}
where $\eta\in\mathbb{R}_{>0}$ is the learning rate. The theoretical minimum of the circular regression loss $-\sum_{i=1}^m\cos\left(y_i-\frac{2\pi}{p}a_is \right)$ is $-m$. We choose a tolerance $\epsilon\in\mathbb{R}_{>0}$ and halt our process once we have
\[-\sum_{i=1}^m\cos\left(y_i-\frac{2\pi}{p}a_is_t \right)\leq -m+\epsilon.\] Our guess for $s$ is then the unique integer in the interval $(s_t-\frac{1}{2}, s_t+\frac{1}{2}]$.


\subsection{Analysis of the algorithm}
\label{subsec:circreg_loss_grad}
\begin{figure}[ht]
    \centering
    \vspace{-4mm}
    \includegraphics[scale=0.65]{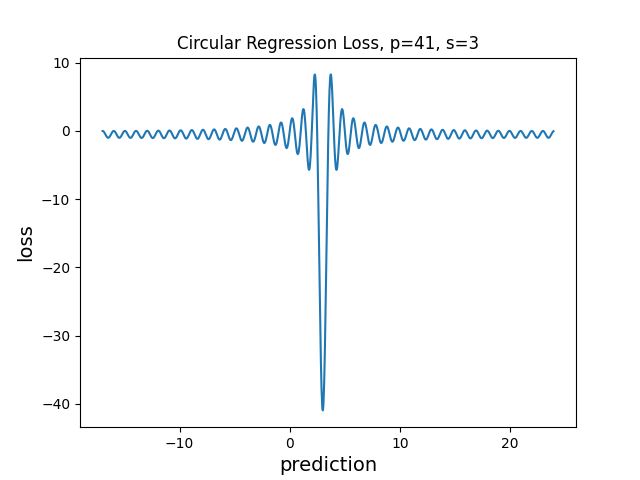}
    \caption{Circular regression loss for $p=41$, $s=3$, plotted using the data set $\{(a_{i}, b_i = a_{i}s \pmod{p})\mid 0\leq a_{i} < p, a_{i}\in \mathbb{Z}\}$, which does not include errors in $b_i$, and has size $m=p$. }
    \label{fig:circregloss}
\end{figure}

First, we visualize the circular regression loss. As shown in Figure \ref{fig:circregloss}, the loss reaches the lowest value $-m$ at $s$, and oscillates with decreasing magnitude as the prediction deviates from $s$. The loss is periodic, hence we only show one interval of length $p$ for simplicity. 
The minimum within an interval of length $p$ is the global minimum. For various values of the prime $p$ and secret $s$, the loss exhibits a shape similar to the plot in Figure \ref{fig:circregloss}(empirically, we observed the loss for $p=23, 41, 71, 113, 251, 367, 967, 1471$). It is highly non-convex, making the search for optima with gradient descent quite challenging. 

Since the loss consists of a deep valley at $s$ but is much closer to 0 everywhere else, we observe that it is reasonable to relax the condition for halting the process. For example, instead of using a tolerance $\epsilon$, we can check the closest integer to the prediction $s_t$ when $-\sum_{i=1}^m\cos\left(y_i-\frac{2\pi}{p}a_is_t \right)\leq -\frac{m}{2}$. 

\begin{figure}[ht]
    \centering
    \vspace{-4mm}
    \includegraphics[scale=0.65]{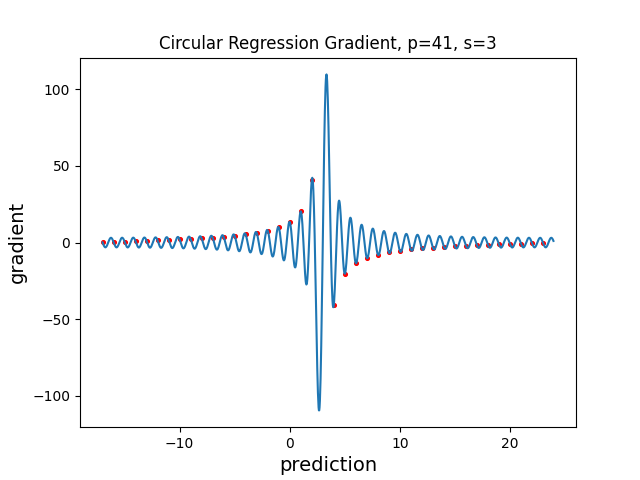}
    \caption{Circular regression gradient for $p=41$, $s=3$, data set $\{(a_{i}, b_i = a_{i}s \pmod{p})\mid 0\leq a_{i} < p, a_{i}\in \mathbb{Z}\}$. The red dots mark the gradient values when the predictions are at integer points. }
    \label{fig:circreggrad}
\end{figure}

For gradient descent to be successful, ideally the direction of the gradient should point toward the closest optimum, and the magnitude of the gradient should be smaller when the prediction is closer to an optimum. Then, as we optimise the prediction using the gradient, we would rapidly approach an optimum and stay close once we reach proximity of an optimum. 
An example of the circular regression gradient is shown in Figure \ref{fig:circreggrad}. 
It oscillates with higher magnitude around $s$, which 
means the magnitude of the gradient displays the opposite behaviour to what we would want in an ideal gradient for gradient descent.
However, the direction of the gradient is good, at least when the predictions are at integer points. When the predictions are at integer points, the gradient at these points (marked with red dots on the plot) always has the sign that points to the closest correct answer.

\begin{figure}[ht]
    \centering
    \vspace{-4mm}
    \includegraphics[scale=0.65]{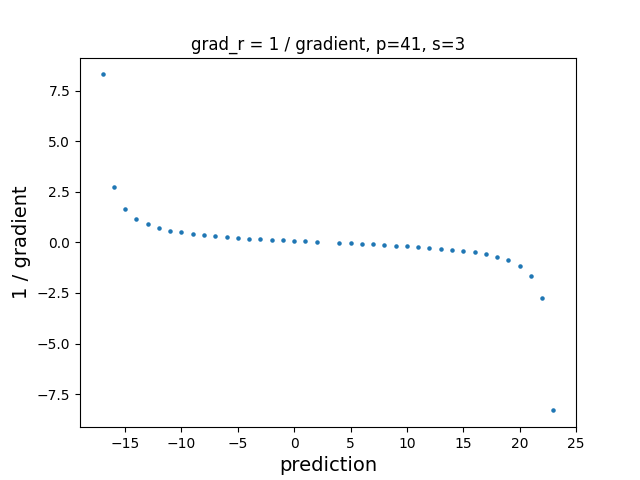}
    \caption{Reciprocal of the circular regression gradient for $p=41$, $s=3$, data set $\{(a_{i}, b_i = a_{i}s \pmod{p})\mid 0\leq a_{i} < p, a_{i}\in \mathbb{Z}\}$, when the predictions are at integer points. }
    \label{fig:circregrecip}
\end{figure}
Since we would like a function with the same sign as the gradient and a magnitude with better behaviour, we are led to consider a version of gradient descent in which we replace the gradient by its reciprocal.
We denote the reciprocal of the gradient by \emph{grad\_r}.
It has the same sign as the gradient and its magnitude has nicer properties, see Figure \ref{fig:circregrecip} as an example, where grad\_r has opposite signs on different sides separated by $s$ and has smaller magnitudes when the prediction is closer to $s$. Grad\_r may explode at various points where the predictions are not integers, where the original gradient is $0$. The gradient is also 0 when the prediction is precisely $s$. 

Note that in practice, we would use a small subset of $\{(a_{i}, a_{i}s \pmod{p})\mid  0\leq a_{i} < p, a_{i}\in \mathbb{Z}\}$ of size $k$ to compute the gradient for efficiency reasons, and the $b_i$ will have errors, which makes the gradient less accurate. Therefore, the grad\_r computed in our implementation does not always have the sign that points to the closest correct answer, nor always having smaller magnitude when the prediction is closer to $s$. 

\subsection{Experiment setup}
\label{subsec:circreg_exp}
We build the data set with vectors $a$ being integers from $1$ to $p-1$, 
and $b = as+e\pmod{p}$, where $e$ has standard deviation $\sigma=3$. For each prime number $p$, we run 20 integer values of $s$, randomly chosen from $1$ to $p-1$ without replacement. 

For regression, we calculate the gradient on batches (with \emph{batch size} $k$) of data randomly chosen from the whole data set, and adjust it by scaling with $\frac{1}{k}$. Essentially, that means instead of taking the summation, we are taking the mean, so that the batch size does not directly affect how much $s_t$ is updated each step. Starting with a random integer $s_0$ as the initial guess, we update the prediction with the reciprocal of the adjusted gradient, scaled by the learning rate $\eta$, as follows: 
$$s_{t+1}=s_t+\eta \left(\frac{2\pi}{p}\frac{1}{k}\sum_{i=1}^k a_i\sin\left(y_i-\frac{2\pi}{p}a_is_t \right)\right)^{-1}.$$

Although we could use the circular regression loss (Section \ref{subsec:circreg_loss_grad}) to evaluate whether our prediction is likely to be correct, in our implementation we instead verify whether $s_t$ matches $s$ by rounding $s_t$ to the nearest integer and checking the magnitude of $as_t-b \pmod{p}$, since it is quite cheap and more reliable. A run terminates if $s_t$ matches $s$, or if the number of steps reaches $p$. The run is successful at step $t$ if it terminates at step $t$ and the prediction $s_t$, rounded to the nearest integer, matches $s$. 
Henceforth, we will refer to this whole process, starting with a random integer $s_0$ and terminating with an output $s_t\in\mathbb{Z}$, as circular regression. Our implementation and code to reproduce visualizations in Section \ref{subsec:circreg_loss_grad} and results in Section \ref{subsec:circreg_results} are available at: \url{https://github.com/meghabyte/mod-math/circ_reg.ipynb}.

\subsection{Empirical results}
\label{subsec:circreg_results}
To choose the parameters, we ran experiments with various learning rates $\eta$ and batch sizes $k$ and counted the number of successes. While $\eta=1$ had more successful trials when $k\in\{256, 512\}$, the performance varied less with $p$ when $\eta=2$, which is desirable (see Table \ref{tab:circreg_params}). Hence, for the following experiments, we use $\eta=2$. And we set the batch size to $k=256$ because it has similar performance with $k=512$, but smaller batch size costs less compute. 

\begin{table}[ht]
\centering
\begin{tabular}{l|ccc|ccc|ccc}
\toprule
$\eta$    & \multicolumn{3}{c}{0.5} & \multicolumn{3}{c}{1} & \multicolumn{3}{c}{2} \\ 
$p$       & 251   & 1471   & 11197  & 251  & 1471  & 11197  & 251  & 1471  & 11197  \\ \midrule
$k=64$  & 6/20 & 0/20 & 0/20 & 16/20 & 12/20 & 4/20 & 15/20 & 17/20 & 11/20  \\
$k=128$ & 14/20 & 4/20 & 0/20 & 15/20 & 19/20 & 6/20 & 14/20 & 8/20 & 11/20  \\
$k=256$ & 18/20 & 11/20 & 4/20 & 19/20 & 17/20 & 15/20 & 16/20 & 13/20 & 14/20  \\
$k=512$ & 17/20 & 15/20 & 11/20 & 20/20 & 15/20 & 18/20 & 14/20 & 14/20 & 15/20  \\ \bottomrule
\end{tabular}
\vspace{3mm}
    \caption{Number of successes in 20 trials for learning rate $\eta = 0.5, 1, 2$ and batch size $k=64, 128, 256, 512$, ran for $p$ of different sizes and $s$ randomly selected from $1$ to $p-1$ (see Section \ref{subsec:circreg_exp}). 
    We upper-bound the batch size with the size of the data set, i.e., for $p=251$ and $k=256, 512$, each batch is the entire data set. }
    \label{tab:circreg_params}
\end{table}

Table \ref{tab:circreg_numsteps} shows the number of steps for successful trials, with batch size $k=256$. As $p$ increases, the success rate remains roughly the same, but the number of steps increases. 
Unfortunately, with batch size $k=256$, the number of steps needed for circular regression to succeed does not consistently give a significant improvement on exhaustive search.
Possible directions for future work could include scaling $\eta$ with $p$, and learning rate decay. 

\begin{table}[ht]
\centering
\begin{tabular}{ll|ll}
\toprule
$p$ & $\log_2 p$ & success & number of steps \\ \midrule
251 & 8 & 16/20 & 1, 1, 1, 1, 2, 11, 24, 28, 37, 44, 62, 109, 118, 171, \\
 & & & 195, 210 \\
1471 & 11 & 13/20 & 121, 199, 213, 234, 306, 324, 371, 488, 507, 699,  \\
 & & & 724, 810, 859 \\
11197 & 14 & 14/20 & 1912, 2294, 2647, 2747, 2799, 3006, 4450, 5349,  \\
 & & & 6277, 7368, 7431, 8104, 8903, 10520 \\
20663 & 15 & 15/20 & 1234, 2759, 3006, 4070, 4288, 4572, 5120, 6117,  \\
 & & & 6517, 9584, 10445, 10846, 11348, 14325, 15542 \\
42899 & 16 & 15/20 & 290, 583, 785, 1098, 3998, 10225, 17005, 18076,  \\
 & & & 19859, 20241, 21553, 22170, 25864, 34798, 35316 \\
115301 & 17 & 12/20 & 10575, 11436, 12805, 15045, 43322, 51372,  \\
 & & & 58295, 69038, 80187, 86451, 104638, 115134 \\
222553 & 18 & 14/20 & 2952, 3048, 3271, 3847, 11959, 17058, 24574, 38624,  \\
 & & & 62084, 73294, 103107, 138868, 160838, 172156 \\
 \bottomrule
\end{tabular}
\vspace{3mm}
    \caption{Number of steps for successful trials. For each value of $p$, we run circular regression on 20 random values of $s$, with $\eta = 2$ and $k=256$. }
    \label{tab:circreg_numsteps}
\end{table}

\section{Transformers for Modular Multiplication}\label{sec:transformer}

 We now move on to an alternative machine learning-based approach to modular multiplication, namely the use of \emph{transformers}, which are a class of deep learning models designed for ``sequence-to-sequence'' tasks: transforming one sequence of elements (e.g. words) to another. 

\subsection{The task}
\label{subsec:transformer_task}
We consider the noiseless version of the task described in Section \ref{circular-task} -- namely, given a data set consisting of $m$ pairs of integers $\{(a_i,b_i)\}_{1\leq i\leq m}$, where $b_i\equiv a_is\pmod{p}$, the task is to find the unknown secret $s$. Knowledge of $s$ together with the ability to perform multiplication modulo $p$ would allow one to take some $a_j$ as an input and generate a valid sample $(a_j,b_j)$ where $b_j\equiv a_js\pmod{p}$. Moreover, being able to reliably predict $b_j$ given $a_j$ would imply knowledge of $s$ (take $a_j=1$).

We train a model $\mathcal{M}$ on the dataset $\{(a_i,b_i)\}_{1\leq i\leq m}$, and determine successful task performance as the model $\mathcal{M}$'s ability to \textit{generalize} to a held-out \textit{test} set of $n$ unknown samples $\{(a_j,b_j)\}_{1\leq j\leq n}$ not seen during training. Truly learning the secret $s$ and modular multiplication would imply perfect accuracy on a held-out test set -- as we demonstrate, we are currently unable to observe this for modular multiplication, suggesting the difficulty of the task. 

Recent works have demonstrated success in training \textit{transformers}, powerful encoder-decoder deep neural networks that are behind some of the best-performing language models (e.g.\ GPT-3), for modular addition \cite{gromov23, nanda23}. These works have demonstrated a surprising phenomenon called \textit{grokking} \cite{grok2}, where training a model for a large number of steps (with 0 training loss) leads to a surprising ``jump'' in generalization capabilities. We therefore consider the transformer architecture as the model class for $\mathcal{M}$, and specifically frame our task as a sequence-to-sequence task: we represent the integer $a_i$  as an input sequence of $t$ tokens $x_{i,1}...x_{i,t}$ in a given base $\mathcal{B}$, and train a transformer-based $\mathcal{M}$ to output $b_i$ represented as an output sequence of $t$ tokens $y_{i,1}...y_{i,t}$ in the same base $\mathcal{B}$. For example, if we use base $10$ then the tokens are the usual digits of an integer written in its decimal representation. We note that in the previously mentioned works, models trained to perform modular addition output a \textit{single} token, and therefore the size of the transformer's vocabulary $|\mathcal{V}|$, or number of possible tokens, is equivalent to the modulus $p$. This is different from our setting, where $\mathcal{M}$ outputs a \textit{sequence} of tokens. In our case, $|\mathcal{V}|$ is equivalent to the base $\mathcal{B}$, and therefore $\mathcal{B}$ influences the overall sequence length that $\mathcal{M}$ needs to generate. Furthermore, the value of the modulus $p$ dictates the total number of input/outputs we can train and evaluate on, so smaller values are more likely to lead to memorization. Indeed, this is what we observe -- see Section~\ref{sec:mem}.

\subsection{Representation and model}
Following \cite{wenger2022salsa}, we train a sequence-to-sequence transformer, varying the number of encoder-decoder layers, but with a fixed model dimension of 512 and 8 attention heads. The vocabulary size is equivalent to the base ($|\mathcal{V}|=\mathcal{B}$) as described above. \textit{Positional encoding} is used to describe the relative positions of tokens in a sequence. Since the order of the digits is of the utmost importance when representing a number, this should be accounted for in our model.
We experiment with two kinds of  positional encodings: fixed sinusoidal, as is standard in language models such as GPT-2, and randomly initialized encodings that are learned over the course of the task, and view optimal representation of position for arithmetic tasks with transformers as an interesting direction for future work.
\begin{figure}[ht]
    \centering
    \includegraphics[scale=0.5]{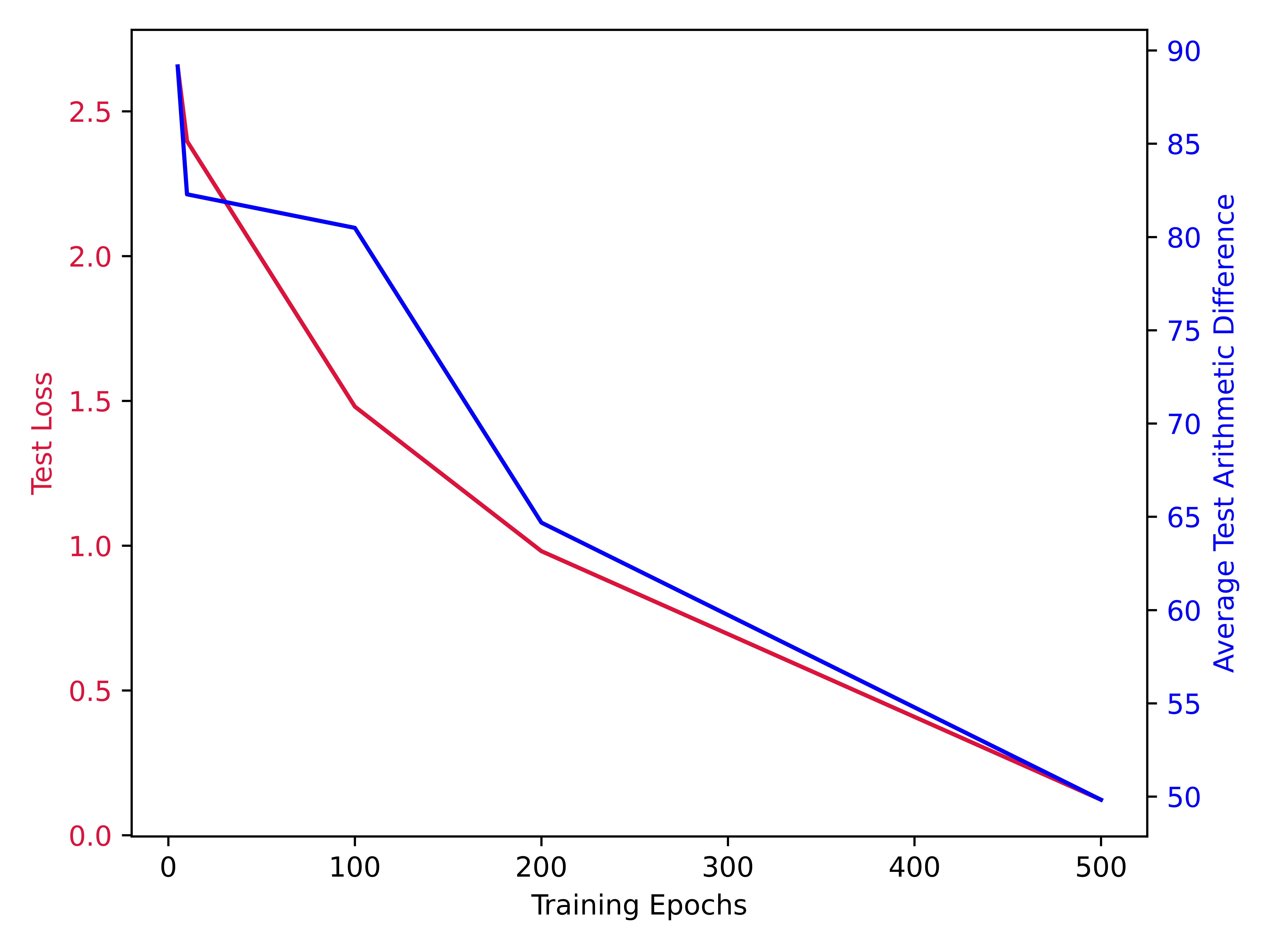}
    \caption{Training curve for modular multiplication task with $p=251$, $s=3$, and base $\mathcal{B}=7$ shows that optimizing sequence-to-sequence accuracy also helps improve arithmetic accuracy, as both test loss and arithmetic difference between generated outputs $\hat{y}_i$ and true values $y_i$ decrease during training.}
    \label{fig:lossvmetric}
\end{figure}
We optimize our model by minimizing the KL-divergence~\cite{KL} between the predicted distribution across all tokens in the vocabulary and the ground truth for each token in the output sequence $y_{i,1}...y_{i,t}$. We also experiment with a \textit{weighted loss} objective that places a higher penalty on divergence in the most significant bits (e.g.\ $y_{i,1}$) than in the least significant bits  (e.g.\ $y_{i,t}$). We specifically use the weight 1.25 for the first $1/3$ most significant bits, $1$ for the middle $1/3$ significant bits, and $0.75$ for the $1/3$ least significant bits. Finally, we implement \textit{early-stopping} by ending training either after $5000$ (the maximum) number of epochs, or when the loss on a held-out valid set has monotonically increased for 5 epochs. Our model implementation and code are publicly available at: \url{https://github.com/meghabyte/mod-math}.

\subsection{Memorization}\label{sec:mem}
\begin{figure}[ht]
    \centering
    \includegraphics[scale=0.5]{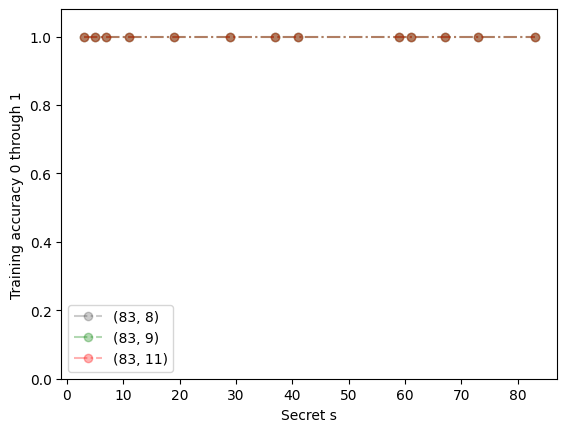}
    \caption{Train accuracy for $p=83$, $3\leq s \leq83$, and $\mathcal{B}\in\{8, 9, 11\}$, after training for $5000$ epochs with learning rate $0.007$.} 
    \label{fig:83line}
\end{figure}

\begin{figure}[ht]
    \centering
    \includegraphics[scale=0.5]{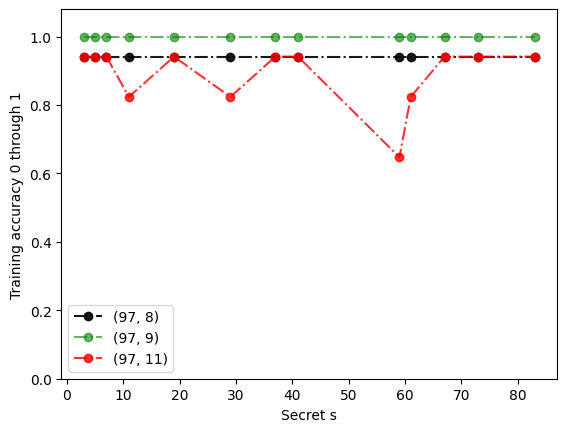}
    \caption{Train accuracy for $p=97$, $3\leq s \leq 97$, and $\mathcal{B}\in\{8, 9, 11\}$, after training for $5000$ epochs with learning rate $0.007$.}
    \label{fig:97line}
\end{figure}

\begin{figure}[ht]
    \centering
    \includegraphics[scale=0.5]{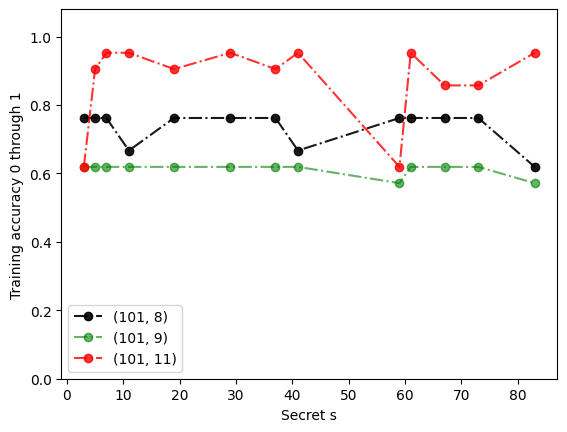}
    \caption{Train accuracy for $p=101$, $3\leq s \leq101$, and $\mathcal{B}\in\{8, 9, 11\}$, after training for $5000$ epochs with learning rate $0.007$.}
    \label{fig:101line}
\end{figure}

\begin{figure}[ht]
    \centering
    \includegraphics[scale=0.5]{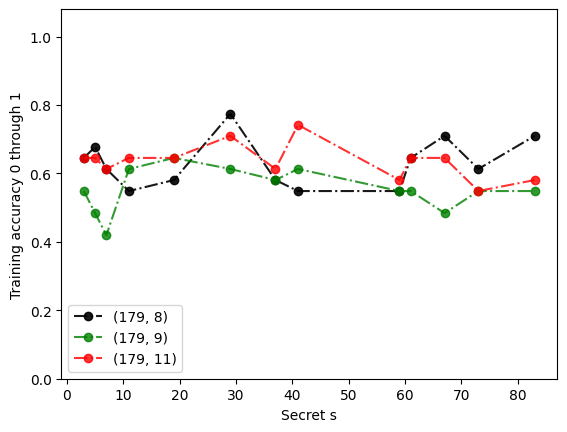}
    \caption{Train accuracy for $p=179$, $3\leq s \leq179$, and $\mathcal{B}\in\{8, 9, 11\}$, after training for $5000$ epochs with learning rate $0.007$.}
    \label{fig:179line}
\end{figure}

\begin{figure}[ht]
    \centering
    \includegraphics[scale=0.5]{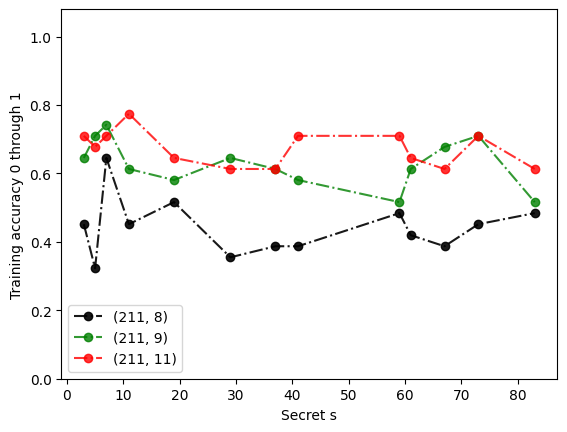}
    \caption{Train accuracy for $p=211$, $3\leq s \leq211$, and $\mathcal{B}\in\{8, 9, 11\}$, after training for $5000$ epochs with learning rate $0.007$.}
    \label{fig:211line}
\end{figure}

\begin{figure}[ht]
    \centering
    \includegraphics[scale=0.5]{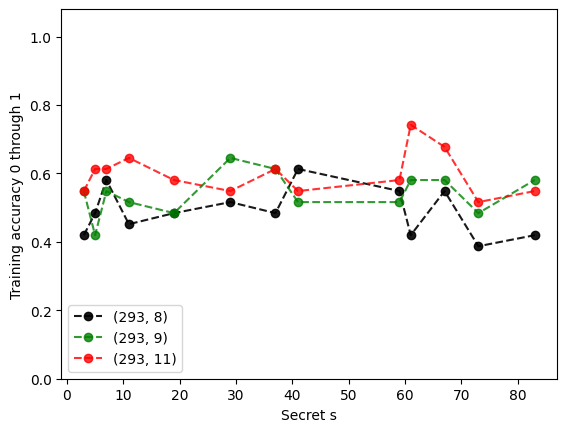}
    \caption{Train accuracy for $p=293$, $3\leq s \leq293$, and $\mathcal{B}\in\{8, 9, 11\}$, after training for $5000$ epochs with learning rate $0.007$.}
    \label{fig:293line}
\end{figure}

We generally observe that for a small prime $p$ and a small secret $s$, the task of memorization results in a high ($\approx 100\%$) model accuracy. This is not entirely surprising as a smaller prime means there are fewer possible inputs and outputs, and therefore there is less to learn.
On the other hand, when the prime $p$ is large and the secret $s$ is small, modular multiplication often coincides with ordinary multiplication. Previous work shows that transformers are capable of learning ordinary multiplication, see e.g.~\cite{Charton, Shen} and the references therein. In our experiments we do not observe increased memorization accuracy when the secret is small compared to the prime, but time constraints mean we have not run experiments with very large primes, so this could be a direction for further investigation. Moreover, memorization may well improve with increased training time. 
In the case of $p=83$, memorization accuracy was  $100\%$ using our current model for base 8, 9 and 11. For $p=97$, memorization accuracy is $100\%$ in base 9, $94.12\%$ for base 8, and varying memorization accuracy for base 11.  
This accuracy quickly decreases for larger primes $p$. We evaluated primes $p$ from $83$ to $293$ for a secret $s$, where $3\leq s \leq 293$ and found that accuracy decreased from $100\%$ to $\approx 40-60\%$ as shown for a selection of primes in Figures \ref{fig:83line} through \ref{fig:293line}. 
We evaluated bases $\mathcal{B}\in\{ 8, 9, 11 \}$ for 5000 epochs with Beam $= 6$ (see Section~\ref{subsec:eval} below for an explanation of beam search). We note that not all bases are equal: in Figure~\ref{fig:97line} we see that for bases $8$ and $9$, memorization is stable and high across all secrets, but for base $11$ we start to see mixed results. This shows that base representation matters when training the model, and this would then influence the model's ability to generalize, cf.~\cite[Section~5.5]{wenger2022salsa}.


\subsection{Evaluation}\label{subsec:eval} In order to evaluate our model's ability to generalize, and therefore successfully learn a given secret $s$, we must first consider the \textit{decoding} method. We experiment with two such methods: \textit{greedy decoding}, where the most likely token $\hat{y}_{i,t}$ is selected  conditioned on the input $x_i$ and previously generated tokens, and \textit{beam search}, where $k$ possible candidates are retained at each step, and the output sequence $\hat{y}$ with the highest likelihood is selected. We then compare the predicted output sequence $\hat{y}_i$ with the true output of modular multiplication for each instance of our test data, $y_i$. 

We evaluate model outputs in two ways: accuracy and arithmetic difference. Consider the task instance $p=251$, $s=3$, and base $\mathcal{B}=7$. For an input $x=426$, the output of modular multiplication would be $y=266$ in base $\mathcal{B}=7$. Perfect accuracy from our model would require $\mathcal{M}$, under a given decoding method, to generate the sequence $\hat{y}_1=2, \hat{y}_2=6, \hat{y}_3=6$. Unfortunately, we largely observe 0$\%$ accuracy on the test set (inputs not observed during training) across different choices of prime modulus, base and secret, and therefore choose to also measure the arithmetic difference (in base 10) between the predicted output $\hat{y}_i$ and ground truth $y_i$. A predicted output sequence of $\hat{y}_1=2, \hat{y}_2=6, \hat{y}_3=3$ would be considered closer in arithmetic difference than $\hat{y}_1=3, \hat{y}_2=6, \hat{y}_3=6$, even though the overall sequence ``loss'' is equivalent (due to a mismatch in one token). A model that can perform modular multiplication under a certain error range in the least significant bits could still be useful for cryptographic attacks. It is not immediately intuitive that optimizing for the sequence-based loss (generating correct tokens) helps decrease the arithmetic difference. However, in Figure \ref{fig:lossvmetric} we show that this generally holds true over a large scale, where both test loss and arithmetic difference decrease as we train $\mathcal{M}$.

\subsection{Generalization results}

\begin{table}[ht]
    \centering
    \begin{tabular}{c|c|c|c|c|c}
    \hline
        ($p$, $s$, $\mathcal{B})$& Std.  & Unweighted Loss  & Sinusoidal Enc.  & Beam = 3 & Beam = 5  \\ \hline
         (97, 11, 9) & 36.5 & 44.475  & 24.613 & 34.975 & 34.975 \\ \hline 
         (101, 3, 7)  & 37.45 & 36.775 & 32.213 & 36.138 & 36.138 \\ \hline
         (109, 29, 8) & 27.288 & 33.913 & 35.425 & 26.963 & 26.963 \\ \hline
         (179, 29, 8) & 68.88 & 75.725 & 83.7 & 57.075 & 57.075 \\ \hline 
    \end{tabular}
    \vspace{3mm}
    \caption{Average test arithmetic difference (lower is better) for different ablations of our transformer modelling approach applied to 3 settings $(p,s, \mathcal{B})$ of modular multiplication. \textbf{Std} refers to training a 2-layer transformer with our weighted loss, random positional encodings, and greedy decoding, with early-stopping implemented up until $5000$ epochs.} 
    \label{tab:compare results}
\end{table}

\begin{figure}[ht]
    \centering
    \includegraphics[scale=0.4]{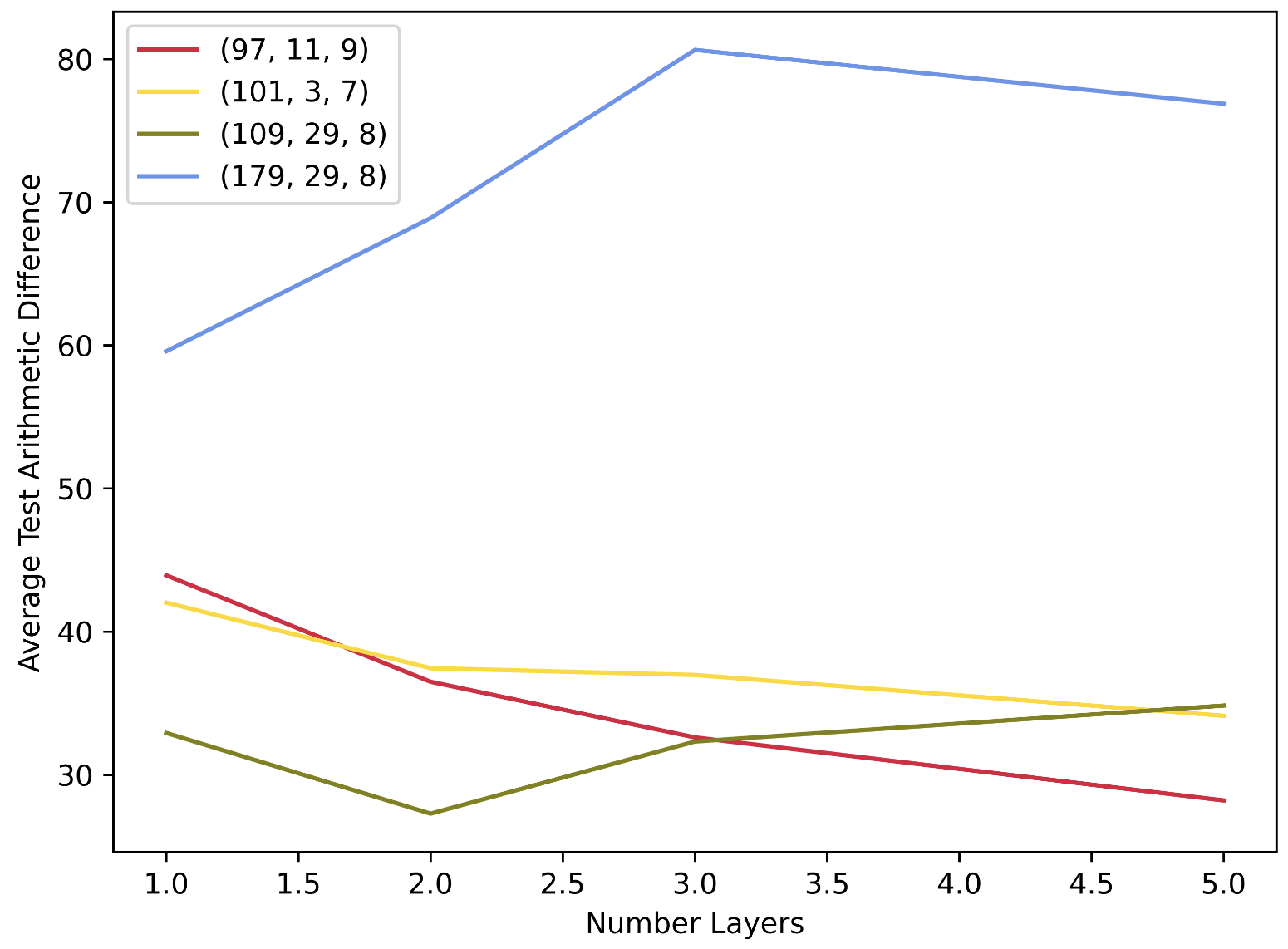}
    \caption{
    Performance vs.\ number of encoder-decoder layers of the model. Legend = ($p, s, \mathcal{B}$). Lower average test arithmetic difference means better performance. }
    \label{fig:layers}
\end{figure}

\begin{figure}
    \centering
    \includegraphics[scale=0.25]{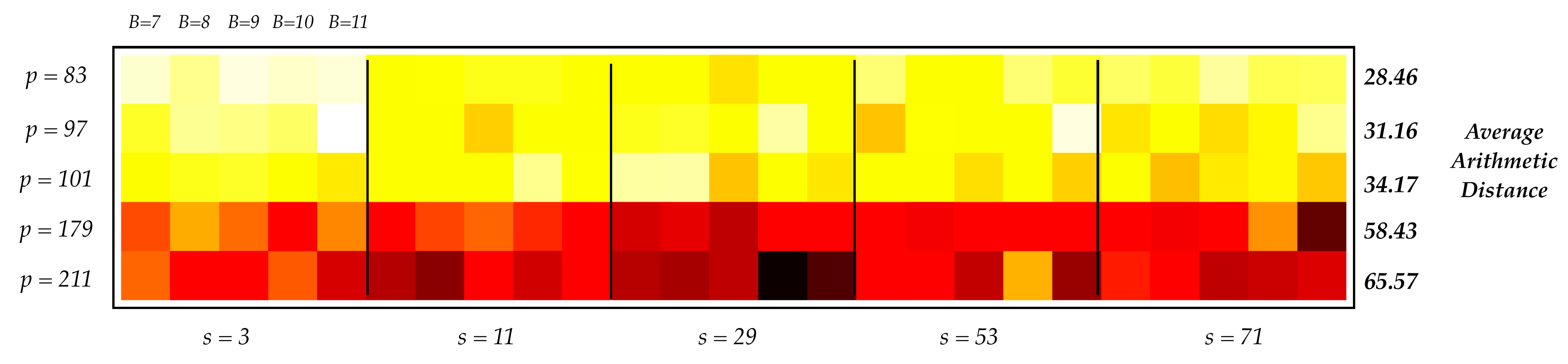}
    \caption{Smaller values of $p$ result in lower average arithmetic difference (error) on held out test examples for our transformers-based approach. Meanwhile, different values of $s$ and the encoding base $\mathcal{B}$ do not show a strong effect on generalization performance. Heatmap colors represent average arithmetic difference with lighter colors meaning smaller values.}
    \label{fig:generalization}
\end{figure}

We generally observe that beam search outperforms greedy decoding for all three instances of $(p,s,\mathcal{B})$, that increasing beam size beyond 3 makes no difference in performance, that sinusoidal position encoding only improves performance over random positional encoding for smaller values of the $p$ modulus, and that weighting the loss generally leads to stronger generalization, see Table~\ref{tab:compare results}. 

Given the low test accuracy (see Section~\ref{subsec:eval}), a natural question to ask is whether this can be improved by having larger models. However, as shown in Figure~\ref{fig:layers}, increasing the number of layers only improves performance for smaller modulus $p$, suggesting that expressiveness of the model is not a sufficient reason for low generalization performance.

Finally, in Figure~\ref{fig:generalization} we show generalization results (average arithmetic distance on held-out test examples) for 5 different values of $p$, 5 different values of $s$, and 5 different values of $\mathcal{B}$, in order to explore the effects of prime, secret, and base. All results are from training models identically, with early-stopping, as described earlier. As expected, smaller values of $p$ result in lower arithmetic difference between the secret $s$ and the generated output, as the space of possible differences is smaller (once normalized by the largest possible difference, which is the value of $p$, all differences lie between $0.3-0.33$). Meanwhile, we observe no trend in performance related to the base $\mathcal{B}$ or the secret $s$, though it is possible that some trend emerges at higher values of $p$.

\section{Discussion}\label{sec:discuss}
One motivation for studying whether modular multiplication is easily tackled by machine learning algorithms, and whether models can learn the representation of a fixed unknown factor (the secret $s$) in multiplication, comes from cryptography. In particular, learning modular multiplication can potentially yield solutions to more advanced problems. For example, consider the problem of recovering a secret $s$ from a data set $\{(a_i, y_i)\}_{1\leq i\leq m}$ where $a_i\in\mathbb{Z}$ and $y_i\equiv g^{a_is}\pmod{p}$ for a public choice of primitive root $g$ modulo the prime $p$. 
This is related to a Diffie--Hellman scheme, where Alice picks a random number $a$ and sends $g^{a}\pmod{p}$ to Bob, 
Bob picks a random number $s$ and sends $g^{s}\pmod{p}$ to Alice, and they both compute $y \equiv g^{as}\pmod{p}$ as the shared secret. Note that Alice does not have access to Bob's random number $s$. However, Alice has control of $a$ and access to the values of $y$ to build a data set for training to predict $y$ from $a$. (Note that this assumes that the secret $s$ belonging to Bob remains fixed, while Alice's $a$ is changing, cf.\ semi-static or ephemeral/static Diffie--Hellman encryption schemes such as ElGamal~\cite{ElGamal}.)
If an algorithm learns to predict $y_i\equiv g^{a_is}\pmod{p}$ from $a_i$, it has implicit knowledge of $s$ and one can potentially extract $s$.

In order to have $(g^{a})^s \equiv g^b \pmod{p}$, we only need $a\cdot s \equiv b\pmod{(p-1)}$, which is a modular multiplication problem. 
Hence, a gradient-based algorithm can try to predict $pred=a\cdot s \pmod{(p-1)}$. However, in the problem described above, the data set has $g^b \pmod{p}$ accessible but $b$ unknown. In fact, solving for $b$ from $y=g^b \pmod{p}$ is itself a famous hard problem known as the discrete logarithm problem. 

Therefore, the loss function would need to involve a comparison with $y = g^b \pmod{p}$. For example, one might try using $g^{pred} \pmod{p} - y$ as the loss function. But this function involves modular arithmetic and raising $g$ to the power of $pred$. Both of these features are challenging for current gradient-based methods, for the following reasons: 
\begin{enumerate}
    \item \label{issue:non-diff} the reduction modulo $p$ function is not differentiable; 
    \item \label{issue:big} $pred$ is an integer of the same scale as $p$ so for large $p$, $g^{pred}$ is a huge number, and the usual ways of handling this (e.g., binary exponentiation) involve modular arithmetic as part of the calculation, which is not differentiable as remarked above. 
\end{enumerate}

To circumvent issue \eqref{issue:non-diff}, one could replace reduction modulo $p$ by a smooth function that gives a close approximation to reduction modulo $p$.
For issue \eqref{issue:big}, writing the numbers in some base $\mathcal{B}$ (see Section \ref{subsec:transformer_task}) could be helpful, illustrated in Example~\ref{Eg:DH} below. 

\begin{example}\label{Eg:DH}
Suppose we would like to train a transformer to predict $pred=a\cdot s \pmod{(p-1)}$ from $a$. The data set consists of pairs $(a,y)$, where $y \equiv (g^{a})^s \pmod{p}$. 
The transformer could be set up to output a sequence that writes $pred$ in base $\mathcal{B}$. Let us denote that sequence as $[y_k, ..., y_1, y_0]$, so $pred = y_k\cdot \mathcal{B}^k + ... + y_1\cdot \mathcal{B} + y_0$, where $0\leq y_i < \mathcal{B}$ for all $i$. 
Let $g_i = g^{\mathcal{B}^i}\pmod{p}$. All the $g_i$ are just constants $<p$. 
Let $mod_p$ denote a smooth function approximating reduction modulo $p$. 
A possible loss function to be minimized could be 
$$
\ g^{pred} \pmod{p} - y \approx  mod_p(g^{pred}) - y
$$
With $pred$ written in base $\mathcal{B}$, this is
\begin{equation*}
    \begin{aligned}
 mod_p\left(g^{\sum_{i=0}^k y_i\cdot \mathcal{B}^i}\right) - y
& =  mod_p\left(\prod_{i=0}^k g_i^{y_i} \cdot ... \cdot g_1^{y_1} \cdot g^{y_0}\right) - y \\
& \approx mod_p\left(\prod_{i=0}^k mod_p(g_i^{y_i})\right) - y
    \end{aligned}
\end{equation*}
If we choose $\mathcal{B}$ to be relatively small, which means all the $y_i$ are relatively small, then the $g_i^{y_i}$ should be reasonable to compute. The terms $mod_p(g_i^{y_i})$ are in the interval $(0,p)$, so their product is memory efficient to compute too. Furthermore, the result is differentiable, and therefore may be amenable to gradient-based methods. 
\end{example}

In addition to modular arithmetic remaining hard to learn for machine learning algorithms, the discreteness and the scale of the numbers used in cryptographic applications also bring engineering challenges.
Example~\ref{Eg:DH} above illustrates one possible way forward to mitigate the difficulties with an algorithmic approach.

\end{document}